\documentclass[lettersize,journal]{IEEEtran}
\usepackage{amsmath,amsfonts}
\usepackage{algorithmic}
\usepackage{algorithm}
\usepackage{array}
\usepackage[caption=false,font=normalsize,labelfont=sf,textfont=sf]{subfig}
\usepackage{textcomp}
\usepackage{stfloats}
\usepackage{url}
\usepackage{verbatim}
\usepackage{graphicx}
\usepackage{cite}
\usepackage[colorlinks,linkcolor=blue,citecolor=blue]{hyperref}
\usepackage{caption}
\usepackage{graphicx}
\usepackage{float} 
\usepackage{color}
\usepackage{colortbl}
\usepackage[table,dvipsnames]{xcolor}
\colorlet{tabfirst}{Green!25}       
\colorlet{tabsecond}{SpringGreen!45} 
\colorlet{tabthird}{Yellow!30}
\usepackage{soul}
\usepackage{booktabs}
\usepackage{multirow}
\usepackage[normalem]{ulem}
\useunder{\uline}{\ul}{}
\usepackage{bbding}

\newcommand{\revise}[1]{\textcolor{black}{#1}}

\begin{document}
\title{Spatially-guided Temporal Aggregation for Robust Event-RGB Optical Flow Estimation}

\author
{
Qianang Zhou,
Junhui Hou,~\IEEEmembership{Senior Member,~IEEE},
Meiyi Yang,
Yongjian Deng,
Youfu Li,~\IEEEmembership{Fellow,~IEEE},
Junlin Xiong,~\IEEEmembership{Member,~IEEE}


\thanks{Qianang Zhou is with the Department of Automation, University of Science and Technology of China, Anhui 230026, China, and is also with the Department of Computer Science, City University of Hong Kong, Hong Kong (email: qianazhou2-c@my.cityu.edu.hk).}

\thanks{Junhui Hou is with the Department of Computer Science, City University of Hong Kong, Hong Kong (email:jh.hou@cityu.edu.hk)}

\thanks{Youfu Li is with the Department of Mechanical Engineering, City University of Hong Kong, Hong Kong (email:meyfli@cityu.edu.hk)}

\thanks{Yongjian Deng is with the College of Computer Science, Beijing University of Technology, Beijing, China (yjdeng@bjut.edu.cn)}

\thanks{Meiyi Yang and Junlin Xiong are with the Department of Automation, University of Science and Technology of China, Anhui 230026, China (email:ymy1996@mail.ustc.edu.cn; xiong77@ustc.edu.cn)}
}

\markboth{Revised Manuscript Submitted to IEEE}%
{Shell \MakeLowercase{\textit{et al.}}: A Sample Article Using IEEEtran.cls for IEEE Journals}


\maketitle

\begin{abstract}
Current optical flow methods exploit the stable appearance of frame (or RGB) data to establish robust correspondences across time. Event cameras, on the other hand, provide high-temporal-resolution motion cues and excel in challenging scenarios. These complementary characteristics underscore the potential of integrating frame and event data for optical flow estimation. However, most cross-modal approaches fail to fully utilize the complementary advantages, relying instead on simply stacking information. This study introduces a novel approach that uses a spatially dense modality to guide the aggregation of the temporally dense event modality, achieving effective cross-modal fusion. Specifically, we propose an event-enhanced frame representation that preserves the rich texture of frames and the basic structure of events. We use the enhanced representation as the guiding modality and employ events to capture temporally dense motion information. The robust motion features derived from the guiding modality direct the aggregation of motion information from events. To further enhance fusion, we propose a transformer-based module that complements sparse event motion features with spatially rich frame information and enhances global information propagation. Additionally, a mix-fusion encoder is designed to extract comprehensive spatiotemporal contextual features from both modalities. Extensive experiments on the MVSEC and DSEC-Flow datasets demonstrate the effectiveness of our framework. Leveraging the complementary strengths of frames and events, our method achieves leading performance on the DSEC-Flow dataset. Compared to the event-only model, frame guidance improves accuracy by 10\%. Furthermore, it outperforms the state-of-the-art fusion-based method with a 4\% accuracy gain and a 45\% reduction in inference time. The code is publicly available at \url{https://github.com/ZhouQianang/STFlow}.
\end{abstract}

\begin{IEEEkeywords}
event-based vision, optical flow, modal fusion.
\end{IEEEkeywords}

\section{Introduction}
\IEEEPARstart{O}{ptical} flow estimation plays a vital role in understanding object motion between image pairs, providing valuable insights into scene dynamics. Frame-based optical flow methods have leveraged the consistent visual appearance of images, leading to extensive development. Over the past decade, learning-based approaches have dominated this field, with correlation-based architectures~\cite{teed2020raft,jiang2021gma} becoming the predominant paradigm. These methods effectively address large-baseline motion by establishing the correlation between all pixel pairs using stable appearance features from frames. However, optical flow estimation faces significant challenges despite its success in general scenarios. Frame data becomes unreliable in high dynamic range or rapid motion environments, where most frame-based algorithms fail to perform effectively~\cite{gallego2020survey}.

\begin{figure}[t]
  \centering
  \includegraphics[width=0.95\linewidth]{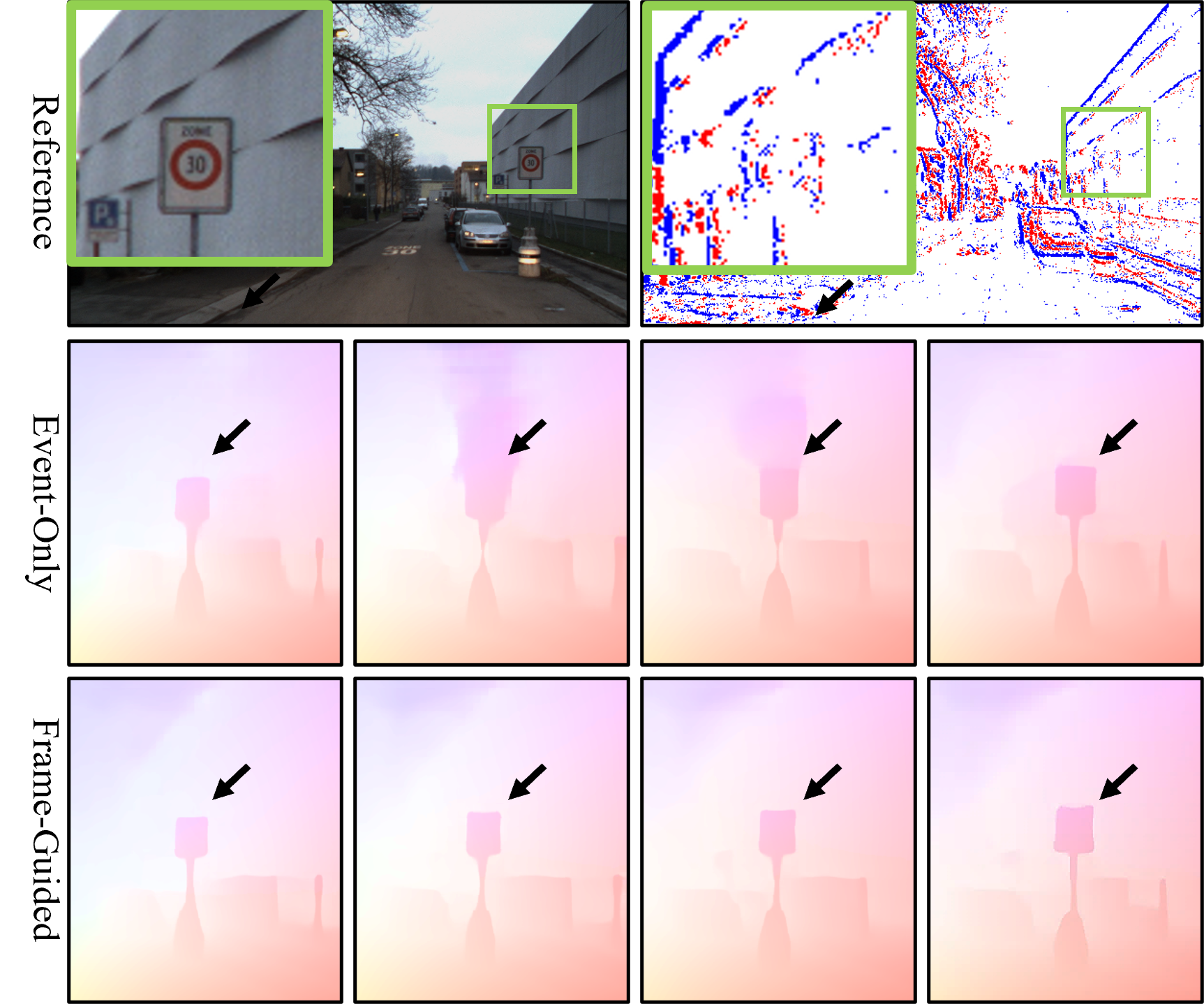}
  \caption{\textbf{Frame-guided prediction improves stability.} The lack of texture in event data leads to unstable predictions. With the proposed strategy that guides the aggregation of temporal features using spatially stable frame, our method achieves more consistent predictions across frames.}
  \label{fig:covercomp}
\end{figure}

Event cameras excel in these challenging scenarios, offering high-frequency motion information that opens new opportunities for visual algorithms~\cite{gallego2020survey,chen2024segment,wu2024motion,ding2023mlb, deng2024dynamic, yao2024sam,jiang2023event, zhu2024continuous}. Their unique characteristics facilitate motion estimation across a broader range of scenes, driving advancements in tasks such as motion deblurring~\cite{xu2021ebmd,chen2024motion, chen2022residual, zhou2023deblurring}, object tracking~\cite{chen2022ecsnetstf,zhu2022learning}, and video interpolation~\cite{paredes2021ebir,liu2024video,ding2024video, zhu2025temporal,yang2024latency}. Recently, event-based optical flow algorithms have gained considerable traction. Unlike frames, event data provides exceptional temporal resolution and dynamic range but lacks a consistent visual appearance. State-of-the-art methods~\cite{liu2023tma,gehrig2024bflow} leverage the high temporal resolution of events to address issues related to spatial sparsity and noise. Although integrating intermediate motion cues has proven to enhance performance, the inherent sparsity of events often results in unstable feature representation. As illustrated in Fig.~\ref{fig:covercomp}, event-only methods face challenges such as a lack of texture, resulting in structural information loss and inconsistent estimates across consecutive frames. These challenges underscore the need for combining frames and events to improve performance.

Given the highly complementary nature of frames and events, various task-specific fusion strategies have been proposed to improve performance~\cite{sun2020ebsegfusion}. However, fusion-based optical flow algorithms remain relatively underdeveloped, particularly in exploiting the complementary strengths of frames and events. For example, Gehrig et al.~\cite{gehrig2024bflow} combine motion features from both modalities through direct concatenation, without accounting for their distinct characteristics. Wan et al.~\cite{wan2022dceiflow} constructs cross-modal correlation maps but overlooks intra-modality similarities. We argue that it is essential to fully account for the unique properties and complementary strengths of different modalities and to design effective fusion strategies specifically tailored for optical flow estimation.

To this end, we analyze the advantages of the two modalities in optical flow estimation. Frame features exhibit a stable visual appearance in most scenarios, enabling robust spatial correspondences. Event features, on the other hand, provide rich temporal information and retain essential visual structures in challenging environments. Moreover, frames are spatially dense but temporally sparse, whereas events exhibit the opposite characteristics. Building on this observation, we propose leveraging frame data to generate robust spatial correspondences and guide the temporal aggregation of event information. On the other hand, we construct temporally dense correlation maps from event data to extract high temporal resolution motion cues. Our approach integrates the spatial stability of frame data with the temporal richness of event data, enabling effective cross-modal fusion for optical flow estimation.

Specifically, we propose a cross-modal collaborative framework, as depicted in Fig.~\ref{fig:framework}. First, we introduce an event-enhanced frame representation to improve the robustness of the guiding modality. We then extract robust guiding motion features from the frame modality and temporally dense motion features from the event modality. To effectively integrate these features, we design a cross-modal aggregation module,  where spatial guiding features supplement sparse event motion features and guide their temporal aggregation. The fused features combine rich spatial information with temporal motion cues, leading to more robust and accurate optical flow predictions. Additionally, we introduce a mix-fusion encoder to extract spatiotemporal context features from both modalities. Our proposed network effectively addresses the instability issues of single-modality methods and achieves substantial improvements over existing fusion approaches, as demonstrated in Fig. \ref{fig:covercomp}. Extensive experiments validate the effectiveness of our framework, which achieves state-of-the-art accuracy on the DSEC-Flow~\cite{gehrig2021eraft}.

In summary, our primary contributions are as follows.

\begin{itemize}
\item \revise{We propose a cross-modal collaboration framework that performs multi-level, modality-specific fusion of frame and event, effectively leveraging their complementary characteristics to enhance optical flow estimation.}
\item \revise{We design a lightweight spatially robust modality and a spatially guided temporal aggregation strategy to facilitate high-level feature fusion between frame and event.}
\item Our method achieves state-of-the-art performance on the DSEC-Flow benchmark and remains competitive with existing supervised approaches on the MVSEC dataset.
\end{itemize}

The rest of this paper is organized as follows: Sec.~\ref{sec:rewo} reviews event and frame-based optical flow and cross-modal fusion. Sec.~\ref{sec:pre} introduces the preliminaries and event representations used in this study. Sec.~\ref{sec:method} details the architecture of our framework. Sec.~\ref{sec:exp} presents and analyzes the experimental results. Finally, Sec.~\ref{sec:conclus} concludes this study.

\section{Related Work}\label{sec:rewo}
\subsection{Learning-based Frame Optical Flow}
Learning-based methods have emerged as the predominant approach in optical flow estimation in recent years. FlowNet~\cite{dosovitskiy2015flownet} proposes end-to-end optical flow estimation, demonstrating the potential of learning-based methods for the optical flow task. Building upon this, PWC-Net~\cite{sun2018pwcnet} introduced a more efficient architecture by integrating pyramid processing, warping, and cost-volume construction. This approach leverages the pyramid concept to capture diverse motion magnitudes, improving both accuracy and efficiency. Lite-FlowNet~\cite{hui2018liteflownet} maintained high accuracy while significantly reducing the model size and inference time, making it more suitable for real-time applications. Despite their advancements, these methods faced limitations in handling large motions due to their localized correlation computation. RAFT~\cite{teed2020raft} tackles this challenge by constructing a 4D correlation volume and leveraging diverse non-correlation features for iterative refinement. More recent developments, such as GMFlow~\cite{xu2022gmflow}, reformulate optical flow as a global matching problem based on the 4D correlation volume. Our method is based on the correlation architecture, constructing spatially robust and temporally dense correlation maps within different modalities.


\begin{figure*}[t]
  \centering
  \includegraphics[width=0.99\linewidth]{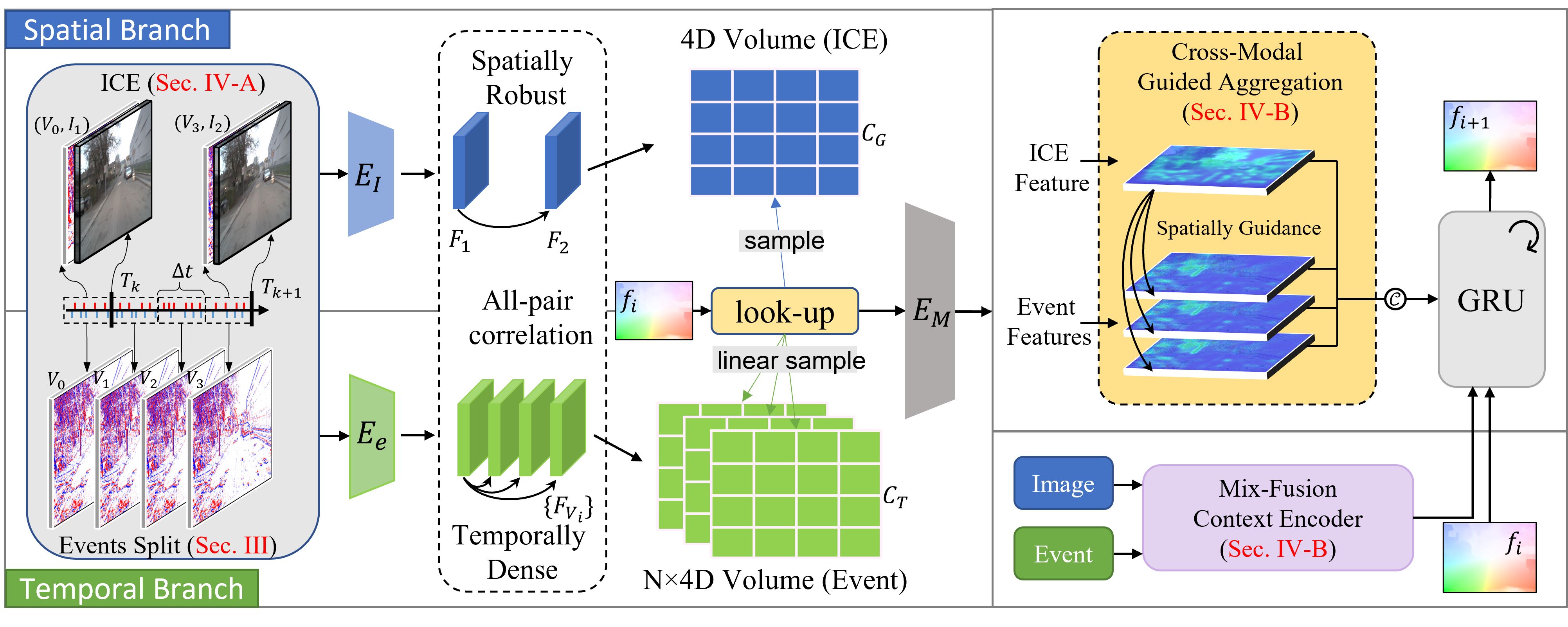}\\
  \includegraphics[width=0.99\linewidth]{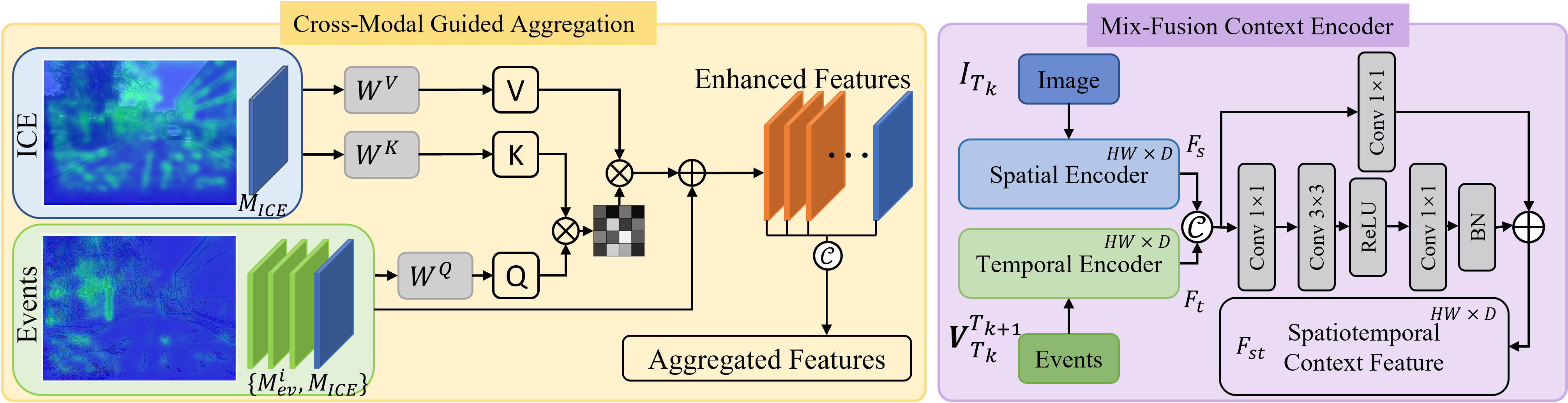}
  \caption{\textbf{The overall architecture of our method}. The core idea of our approach is to leverage the spatial stability of the frame modality to guide the temporal aggregation of event features, enabling complementary interactions between modalities. To achieve this, an event-enhanced frame representation ICE is introduced as robust spatial guidance. The ICE feature is then used as the $K$ to guide the aggregation of temporally dense event features and also serves as the $V$ to enhance the spatially sparse events. Furthermore, contextual features from both modalities are fused to enhance optical flow estimation.}
  \label{fig:framework}
\end{figure*}

\subsection{Event-based Optical Flow} 
Event-based optical flow has seen significant advancements in the past decade. Early approaches~\cite{mueggler2015fitplane} fit local planes in the event stream to estimate optical flow, while others~\cite{pan2020vaof,bardow2016vaof} minimized an energy function using variational optimization. Shiba et al.~\cite{shiba2022multicm} warps events along point trajectories, deriving an intuitive flow solution. 
They further extended this optimization framework to a broader range of event-based vision tasks in~\cite{shiba2024secrets}.
Recently, learning-based methods have demonstrated distinct advantages. Self-supervised methods~\cite{zhu2018evflownet,hagenaars2021nips} primarily use contrast maximization~\cite{gallego2018cm} or temporal loss~\cite{zhu2018evflownet} to supervise the learning process, while supervised methods rely on labeled data. Notable examples include EvFlowNet~\cite{zhu2018evflownet} and E-RAFT~\cite{gehrig2021eraft}, which successfully adapted FlowNet~\cite{dosovitskiy2015flownet} and RAFT~\cite{teed2020raft} architectures for event data. Based on E-RAFT, TMA~\cite{liu2023tma} further utilizes the high temporal resolution of event data to compensate for spatial sparsity. Recently, TCM~\cite{paredes2023taming} introduced multi-scale timestamp loss to supervise dense optical flow prediction, achieving notable success. Moreover, Wan et al.~\cite{wan2022dceiflow} and Gehrig et al.~\cite{gehrig2024bflow} integrated image and event data to estimate dense optical flow. 
Our method emphasizes the use of the unique characteristics of each modality and their complementary advantages to enable effective cross-modal collaboration.

\subsection{Event-Frame Cross-Modal Fusion} 
The complementary nature of events and frames has been extensively explored in various tasks such as semantic segmentation~\cite{sun2020ebsegfusion,zhang2023ebsegfusion,xie2024eisnet,jiang2023evplug}, depth estimation~\cite{mostafavi2021ebdepfusion,gehrig2021ebdepth,cho2022ebdepth}, and object tracking~\cite{zhu2023cross,chen2024crossei}. In these applications, it is common to employ a multi-stage fusion of event and image features within a UNet-like framework, followed by processing the fused features through the decoder. Chen et al.~\cite{chen2024crossei} develop an adaptive sampling method to align event and image modalities, along with a bidirectional-enhanced framework to facilitate cross-modal tracking. Zhu et al.~\cite{zhu2023cross} introduce a mask modeling strategy to promote proactive interaction between tokens from different modalities. Several approaches have also explored the integration of events and frames for optical flow estimation. For example, DCEIFlow~\cite{wan2022dceiflow} generates a pseudo-feature for the second frame by fusing the first frame with events and then estimates the optical flow based on the RAFT structure. BFlow~\cite{gehrig2024bflow} concatenates motion features from both images and events into a single representation and then estimates the control point location of the trajectory. Our method guides the aggregation of temporal motion features using spatially robust frame motion features, enabling the fusion of motion cues.

\section{Preliminary and Data Representation}\label{sec:pre}

The event and frame data are significantly different, and it is necessary to reorganize them. An event is triggered when a pixel of the event camera detects a change in luminance above a threshold $C$. Each event $e_i$ typically includes the time $t_i$, coordinates $(x_i,~y_i)$ and the polarity $p_i$ of its occurrence. The asynchronous events are usually converted into a frame-like representation. In this paper, we convert the event set into a voxel grid $\mathbf{V}$, as in the previous works~\cite{zhu2018evflownet}:
\begin{equation}
\begin{split}
t_i^* &= (B-1)(t_i-t_1)/(t_N-t_1),\\
\mathbf{V}(x,y,t) &= \sum\limits_{i}p_ik_b(x-x_i)k_b(y-y_i)k_b(t-t_i^*),\\
k_b(a)&=\max(0,1-|a|),
\end{split}
\label{eq:voxel}
\end{equation}
where $B$ represents the number of time bins, $t_i^*$ discretizes $t_i$ to the $i$-th time bin, and $k_b(a)$ is a bilinear interpolation function.

Different time ranges of events and frame data are required to estimate pixel motion from $T_k$ to $T_{k+1}$. For frames data, we use images $I_k$ and $I_{k+1}$, corresponding to timestamps $T_k$ and $T_{k+1}$. The event data representation is illustrated in Fig.~\ref{fig:framework}. We capture temporally dense motion cues by uniformly partitioning the event stream within the interval \([T_k, T_{k+1}]\) into a sequence of target segments $\{\mathbf{V}_1, \mathbf{V}_2, \ldots, \mathbf{V}_N\}$, each with an average time span of $\Delta t$. Then we use the segment from the interval $[T_k - \Delta t, T_k]$ as the reference segment $\mathbf{V}_0$. Correlations between the reference and targets provide rich temporal motion information for flow estimation.


\section{Proposed Method}\label{sec:method}
To facilitate cross-modal collaboration in optical flow estimation, we leverage spatially robust features to guide the temporal aggregation of sparse event features. Sec.~\ref{sec:m1} introduces an event-enhanced frame representation to improve the robustness of guidance, and discusses spatial and temporal motion feature extraction. Sec.~\ref{sec:m2} details how robust spatial features guide the aggregation of temporally dense event features and our proposed spatiotemporal context feature.

\subsection{Spatial Guidance and Temporal Features Construction.}\label{sec:m1}
As shown in Fig.~\ref{fig:framework}, we extract spatially robust correlation features in the spatial branch and temporally dense motion cues in the temporal branch. Building on the data representations outlined in Sec.~\ref{sec:pre}, each branch is detailed as follows.

\begin{figure}[t]
  \centering
  \includegraphics[width=0.90\linewidth]{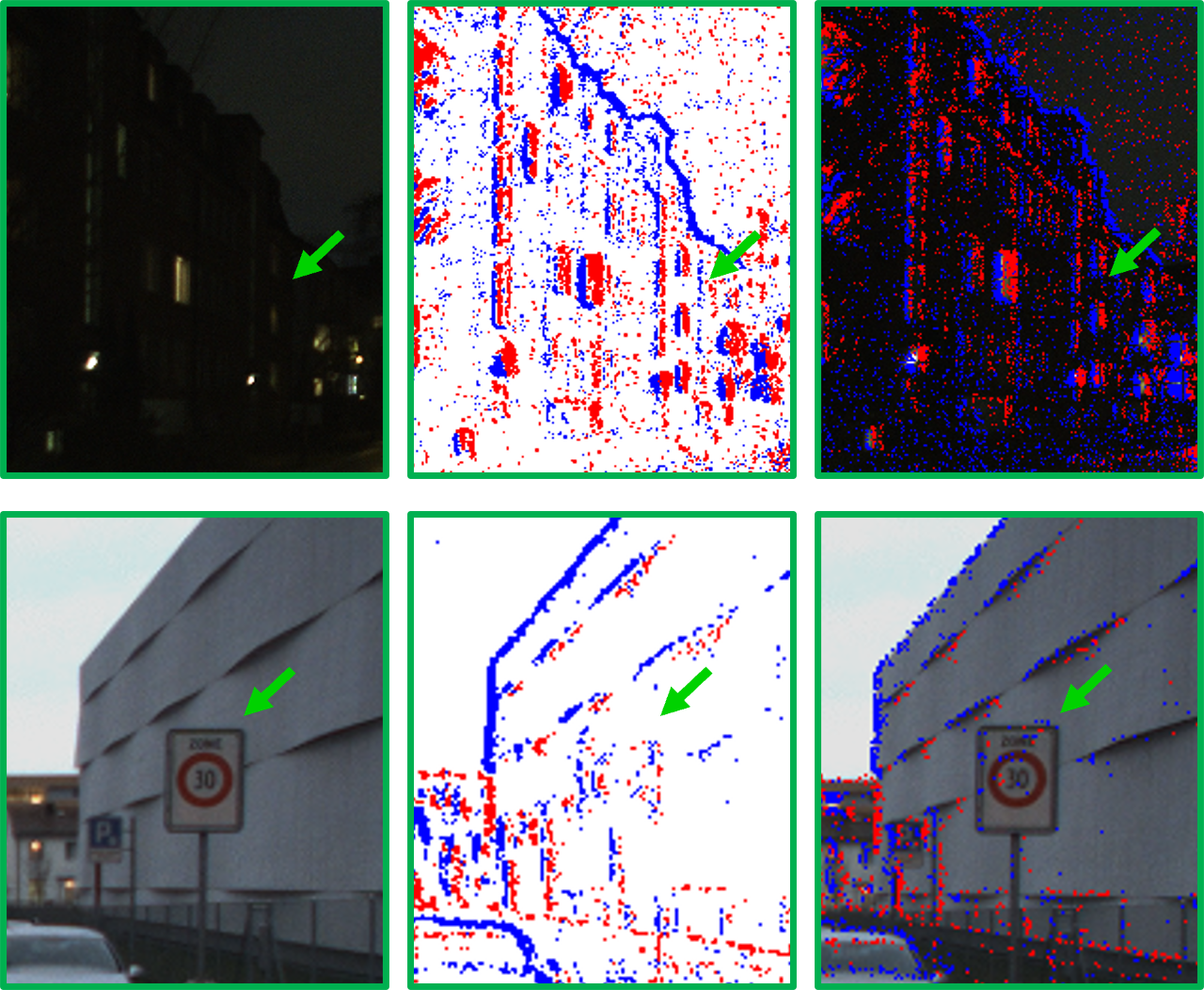}
  \caption{\textbf{Complementary strengths of frames and events across scenes.} Frames provide rich texture under normal lighting but often fail in challenging illumination, where events preserve clear structural information.}
  \label{fig:ice}
\end{figure}

\vspace{0.5em}
\textbf{Robust Spatial Guidance.}
Correlation-based methods establish reliable pixel correspondences between two viewpoints to extract motion features. However, high-temporal-resolution event data are often sparse and noisy, complicating accurate optical flow estimation. To address this, we use robust frame features to guide the aggregation of temporally dense event motion features. Frames are susceptible to disturbances like illumination changes and motion blur, so we enhance frame data with events to improve guidance stability. Although events are typically used to enhance frame quality, we introduce a simplified representation called Image-Event Connection (ICE) that avoids complex networks~\cite{paredes2021ebir} for reconstructing high-quality images. As illustrated in Fig.~\ref{fig:ice}, ICE systematically organizes events and frames, ensuring robust guidance across diverse environments.


Specifically, given the input event voxel $\mathbf{V}$ and image $\mathbf{I}$, we first mapped pixel values and voxel values to $[-1,1]$ to balance the contributions of events and frames:
\begin{equation}
\begin{split}
    \hat{\mathbf{I}} &= 2\times\frac{\mathbf{I}}{255}-1,\\
    \hat{\mathbf{V}} &= \frac{\mathbf{V}}{\max{(|\mathbf{V}|)}+\epsilon},
\end{split}
\end{equation}
\revise{where $\epsilon$ is a small positive constant to prevent the maximum value from being zero, and $\max(|\mathbf{V}|)$ is computed individually for each sample.} The ICE is then generated as: 
\begin{equation}
\mathbf{ICE}=\texttt{concat}(\hat{\mathbf{V}}_{t-\Delta t}^{t},\hat{\mathbf{I}}_t),
\end{equation}
where $\mathbf{I}_t$ corresponds to time $t$, and $\mathbf{V}_{t-\Delta t}^t$ represents the voxel from $t-\Delta t$ to $t$.

As shown in Fig.~\ref{fig:ice}, events and frames exhibit strong complementarity.
In low-light conditions, frame data may degrade significantly, \revise{while event data still provides sufficient spatial information}. In well-lit scenes, frame intensity information complements the event data. Consequently, the motion features extracted from ICE exhibit high spatial stability, ensuring robust guidance. Note that we use event only within a time window of length $\Delta t$ to avoid edge blurring caused by long windows.
The event time window of ICE matches the temporal length in the event target. This design maintains consistency during cross-modal aggregation while avoiding unnecessary computational overhead.

Subsequently, we extract features from ICE pairs and generate all-pairs correlation volume $\mathbf{C}_{G}$ between them for guidance:
\begin{equation}
\mathbf{C}_{G}=\frac{\mathbf{F}_1\mathbf{F}_2}{\sqrt{D}},
\end{equation}
\revise{where $\mathbf{F}_1, \mathbf{F}_2$ denote the features corresponding to the start and end ICEs}, and $D$ is the feature channels. We emphasize that $\mathbf{C}_G\in \mathbb{R}^{HW\times HW}$ encompasses not only the motion information for the entire optical flow from $T_k$ to $T_{k+1}$ but also the robust spatial correlation generated from the ICE, thereby supporting the guidance for event features.


\vspace{0.5em}
\noindent\textbf{Temporally Dense Correlation.} 
The effectiveness of leveraging the high temporal resolution of event data to compensate for sparsity and noise has been validated in several recent works. Notably, TMA~\cite{liu2023tma} and BFlow~\cite{gehrig2024bflow} improve performance by replacing the single correlation volume with temporally-dense correlation volumes. We believe this design benefits from the temporal consistency of structural motion patterns, while event noise, especially in low-light conditions, is generally uncorrelated across time. Aggregating correlation information across multiple time steps thus enhances the signal-to-noise ratio by reinforcing consistent features and suppressing transient noise.
Building on this insight, we extract event motion cues at multiple intermediate times, following these works.

Specifically, we first split the event stream into a reference $\mathbf{V}_0$ and a series of targets $\{\mathbf{V}_n,n\in[1,N]\}$, as described in Sec.~\ref{sec:pre}. The reference $\mathbf{V}_0$ and the final target $\mathbf{V}_N$ are aligned with the first and second frames $I_k$ and $I_{k+1}$, respectively. To ensure consistency, all the targets and the reference share the same feature encoder. The correlation volumes are then generated between the reference and each target to construct the temporally dense cost volumes:
\begin{equation}
    \mathbf{C}_{T} =\{ \frac{\mathbf{F}_{\mathbf{V}_0}\mathbf{F}_{\mathbf{V}_n}}{\sqrt{D}}\},n\in[1,N],
\end{equation}
where $\mathbf{F}_{\mathbf{V_n}}$ is the $n$-th feature of events and $D$ is the feature channels.

Cost volumes $\mathbf{C}_T$ contain motion information with high temporal resolution, yet they are spatially sparse and unstable. Previous work~\cite{gehrig2024bflow} adopted an overlapping strategy to mitigate the adverse effects of sparsity, which sacrifices efficiency and becomes unnecessary with the introduction of frames. The features in $\mathbf{C}_T$ lack full optical flow information and scene texture, which are precisely what $\mathbf{C}_{G}$ contains. Consequently, the motion feature extracted from $\mathbf{C}_{G}$ will be used to guide the aggregation of temporal event features from $\mathbf{C}_T$ in the following.

\subsection{Context Fusion and Guided Aggregation}\label{sec:m2}

\vspace{0.5em}
\textbf{Spatiotemporal Context.}
The context encoder extracts features from event or image data to guide optical flow estimation. Spatial context features from frames capture the spatial structural details of $I_k$, while temporal context features from events encode information spanning from $T_k$ to $T_{k+1}$. Previous studies have shown that both features enhance optical flow prediction~\cite{gehrig2021eraft,liu2023tma,gehrig2024bflow,wan2022dceiflow}. However, the interplay between spatial and temporal features, as well as their individual contributions, remains unexplored. We argue that combining spatial and temporal contexts yields richer, more informative representations. This paper introduces spatiotemporal context features as a replacement for single-modal features, with further analysis presented in the experimental section.

To achieve this, we propose the Mix-Fusion Context Encoder, illustrated in Fig.~\ref{fig:framework}, to extract spatiotemporal context features from events and frames. \revise{We first extract spatial context features from $I_k$ and temporal context features from the event voxel between $T_k$ and $T_{k+1}$, using two separate encoders.} Both features are designed to have the same channel dimensions and are subsequently fed into the mix-fusion block. The mix-fusion module is inspired by~\cite{xie2021segformer}, and the output spatiotemporal feature $\mathbf{F}_{st}$ has the same number of channels as single-modal context features:
\begin{equation}
\begin{split}
\mathbf{H}&=\texttt{concat}(\mathbf{F}_{s},\mathbf{F}_{t}),\\
\mathbf{F}_{st} &= \texttt{MLP}(\texttt{Conv}_{3\times 3}(\texttt{MLP}(\mathbf{H}))+ \texttt{MLP}(\mathbf{H}),
\end{split}
\end{equation}
where $\texttt{MLP}(\cdot)$ performs per-pixel feature fusion, and $\texttt{Conv}_{3\times 3}(\cdot)$ is used for local information propagation.

\vspace{0.5em}
\noindent\textbf{Cross-Modal Guided Aggregation.}
We have generated spatially robust ICE correlation volumes and temporally dense event correlation volumes. In the update branch, ICE motion features guide the temporal aggregation of event features, iteratively refining the optical flow estimation with the aid of spatiotemporal context features.


Correlation-based methods sample the cost map corresponding to the current estimation from the correlation volume, encoding it as motion features to drive further refinement. Our cross-modal approach samples from both modalities using the lookup operation. For the ICE modality, given the current optical flow estimate $\hat{f}$, the cost map of $\hat{f}$ is sampled as:
\[Cost = \texttt{lookup}(\mathbf{C}_G,\hat{f}),\]
where $\texttt{lookup}(\cdot,\cdot)$ samples the similarity between pixel pairs from $\mathbf{C}_G$ based on $\hat{f}$. For the event modality, each reference-target pair has different time intervals, necessitating a linear lookup strategy~\cite{liu2023tma}. Given the flow estimation $\hat{f}$, the $i$-th cost map is sampled as:
\[Cost=\texttt{lookup}(\mathbf{C}_T^i,\hat{f}\frac{i}{N}), i\in[1,N], \]
where $\mathbf{C}_T^i$ is the $i$-th correlation maps in $\mathbf{C}_T$. 

The sampled cost maps are encoded as motion features by $E_M$, as shown in Fig.~\ref{fig:framework}. The ICE motion feature $\mathbf{M}_{ice}$ encodes robust visual information, while the event motion features $\{\mathbf{M}_{ev}^i,i\in[1,N]\}$ provide temporally detailed motion cues. Subsequently, we use $\mathbf{M}_{ice}$ to guide the aggregation of $\mathbf{M}_{ev}$. 
\revise{Inspired by~\cite{jiang2021gma,liu2023tma}, we adopt a lightweight Transformer to implement the cross-modal guided aggregation. Event motion features $\mathbf{M}_{ev}$ are used as \textbf{queries}, while ICE motion features $\mathbf{M}_{ice}$ serve as both \textbf{keys} and \textbf{values}, guiding and facilitating the event features through token-level cross-attention:}
\begin{equation}
\begin{split}
\mathbf{Q}_{ev}^i = \mathbf{M}_{ev}^{i}&W^Q,~\mathbf{Q}_{img}=\mathbf{M}_{ice}W^Q,\\
\mathbf{K}&=\mathbf{M}_{ice}W^K,\\
\mathbf{V}&=\mathbf{M}_{ice}W^V,\\
\mathbf{AM} = \mathbf{M} + &\texttt{ffn}(\texttt{softmax}(\frac{\mathbf{QK}}{\sqrt{D}})\mathbf{V}),
\end{split}
\end{equation}
where $D$ is the dimension of $\mathbf{K}$, and $\texttt{ffn}(\cdot)$ is the feed-forward neural network. 
\revise{In our experiments, we use a single-head, single-layer configuration with shared embedding dimensions.}
Eventually, the aggregated features are concatenated into a single motion feature, which is fed into the $\texttt{ConGRU}$ for further refinement.

\subsection{Supervision}\label{sec:mC}
We follow the standard setup of correlation-based methods to supervise the network output. The $L_1$ distance between the predictions and the ground truth is taken as the loss, and the supervision is performed on each output of the iterator:
\[\mathcal{L}=\sum\limits_{j=1}^{n}\gamma^{n-j}\|\hat{f}_j-f_{gt}\|_1,\]
where $n$ is the total number of \texttt{ConvGRU} iterations, $\hat{f}_j$ is the output of the $j$-th iteration, and $\gamma$ is the decay factor.

\section{Experiments}\label{sec:exp}

\begin{table*}[t]
\centering
\caption{\textbf{Detail results on DSEC-Flow}. Best results are highlighted as \colorbox{tabfirst}{\bf first}, and \colorbox{tabsecond}{second}. `E' represents events, and `I' represents images. ``$\downarrow$" indicates the smaller, the better. All results are available on the public online benchmark of DSEC-Flow.}
\begin{tabular}{@{}lccccccccccccc@{}}
\toprule \toprule
\multirow{2}{*}{Method} & \multirow{2}{*}{Input} & \multicolumn{3}{c}{Overall}                      & \multicolumn{3}{c}{interlaken\_00\_b}            & \multicolumn{3}{c}{interlaken\_01\_a}            & \multicolumn{3}{c}{thun\_01\_a}                  \\ \cmidrule(l){3-14} 
                      &  & EPE($\downarrow$)            & 3PE($\downarrow$)            & AE($\downarrow$)             & EPE($\downarrow$)            & 3PE($\downarrow$)            & AE($\downarrow$)             & EPE($\downarrow$)            & 3PE($\downarrow$)            & AE($\downarrow$)             & EPE($\downarrow$)            & 3PE($\downarrow$)            & AE($\downarrow$)             \\ \midrule
ERAFT~\cite{gehrig2021eraft}        & E           & 0.79           & 2.68           & 2.85           & 1.39          & 6.19          & 2.36          & 0.90          & 3.91          & 2.54          & 0.65          & 1.87           & 2.94          \\
TMA~\cite{liu2023tma}               & E           & 0.74           & 2.30           & 2.68           & 1.39          & 5.79          & 2.157          & 0.81          & 3.11          & 2.23          & 0.62          & 1.61          & 2.88           \\
IDNet~\cite{wu2024idnet}            & E           & 0.70           & \cellcolor{tabsecond}{1.96}           & 2.58           & 1.25           & \cellcolor{tabsecond}{4.35}          & 2.11           & 0.77          & 2.60          & 2.25           & 0.57          & 1.47          & 2.66          \\
ECDDP~\cite{yang2023ecddp}          & E           & 0.72           & 2.04           & 2.72           & 1.31           & 5.03          & 2.00          & \cellcolor{tabsecond}{0.76}          & \cellcolor{tabsecond}{2.47}          & 2.18          & \cellcolor{tabfirst}{0.52}  & \cellcolor{tabsecond}{1.36}          & \cellcolor{tabfirst}{2.34} \\ \midrule
BFlow~\cite{gehrig2024bflow}        & E+I           & \cellcolor{tabsecond}{0.69}           & 2.02           & \cellcolor{tabsecond}{2.42}           & \cellcolor{tabfirst}{1.11}           & 4.58          & \cellcolor{tabfirst}{1.93}          & 0.77          & 2.63          & \cellcolor{tabsecond}{2.07}          & 0.59  & 1.59          & 2.71 \\
\textbf{Ours}              & E+I    & \cellcolor{tabfirst}{0.66}  & \cellcolor{tabfirst}{1.68}  & \cellcolor{tabfirst}{2.37}  & \cellcolor{tabsecond}{1.22} & \cellcolor{tabfirst}{4.32} & \cellcolor{tabsecond}{1.96} & \cellcolor{tabfirst}{0.72} & \cellcolor{tabfirst}{2.10} & \cellcolor{tabfirst}{2.06} & \cellcolor{tabsecond}{0.55}          & \cellcolor{tabfirst}{1.31} & \cellcolor{tabsecond}{2.43}          \\ \midrule
\multirow{2}{*}{Method} & \multirow{2}{*}{Input} & \multicolumn{3}{c}{thun\_01\_b}                  & \multicolumn{3}{c}{zurich\_city\_12\_a}          & \multicolumn{3}{c}{zurich\_city\_14\_c}          & \multicolumn{3}{c}{zurich\_city\_15\_a}          \\ \cmidrule(l){3-14} 
                    &    & EPE($\downarrow$)            & 3PE($\downarrow$)            & AE($\downarrow$)             & EPE($\downarrow$)            & 3PE($\downarrow$)            & AE($\downarrow$)             & EPE($\downarrow$)            & 3PE($\downarrow$)            & AE($\downarrow$)             & EPE($\downarrow$)            & 3PE($\downarrow$)            & AE($\downarrow$)             \\ \midrule
ERAFT~\cite{gehrig2021eraft}   & E                & 0.58          & 1.52          & 2.20          & 0.61          & 1.06          & 4.50          & 0.71          & 1.91          & 3.43          & 0.59          & 1.30          & 2.55          \\
TMA~\cite{liu2023tma}          & E           & 0.55          & 1.31           & 2.10          & 0.57          & 0.87          & 4.38          & 0.66          & 1.99           & 3.09          & 0.55          & 1.08          & 2.51          \\
IDNet~\cite{wu2024idnet}       & E            & 0.55          & 1.35           & 2.07           & 0.60          & 1.16          & 4.56          & 0.76           & 2.74          & 3.74          & 0.55          & 1.02          & 2.55          \\
ECDDP~\cite{yang2023ecddp}     & E              & \cellcolor{tabsecond}{0.51}          & \cellcolor{tabsecond}{1.21}          & 1.93           & 0.55          & 0.76          & 4.35          & 0.69          & 2.39          & 3.22          & \cellcolor{tabfirst}{0.52} & \cellcolor{tabfirst}{0.89} & 2.41          \\ \midrule
BFlow~\cite{gehrig2024bflow}    & E+I               & 0.55           & 1.42           & \cellcolor{tabsecond}{1.84}           & \cellcolor{tabsecond}{0.54}           & \cellcolor{tabsecond}{0.72}          & \cellcolor{tabsecond}{3.86}          & \cellcolor{tabsecond}{0.59}          & \cellcolor{tabsecond}{1.37}          & \cellcolor{tabfirst}{2.44}          & 0.57  & 1.24          & \cellcolor{tabsecond}{2.38} \\
\textbf{Ours}              & E+I    & \cellcolor{tabfirst}{0.50} & \cellcolor{tabfirst}{1.12} & \cellcolor{tabfirst}{1.76} & \cellcolor{tabfirst}{0.50} & \cellcolor{tabfirst}{0.54} & \cellcolor{tabfirst}{3.80} & \cellcolor{tabfirst}{0.55} & \cellcolor{tabfirst}{0.94}  & \cellcolor{tabsecond}{2.47} & \cellcolor{tabsecond}{0.53}          & \cellcolor{tabsecond}{0.92}          & \cellcolor{tabfirst}{2.27} \\ \bottomrule \bottomrule
\end{tabular}\label{tab:dsecall}
\end{table*}

\subsection{Datasets and Setup}
To ensure fair comparisons with prior methods, we conducted extensive experiments on MVSEC~\cite{zhu2018mvsec} and DSEC-Flow~\cite{gehrig2021dsec}. Both datasets provide frame and event data from real-world scenes and are widely adopted for evaluating event-based optical flow. MVSEC includes outdoor driving sequences and several indoor flying sequences. We followed the same setup as in prior work~\cite{zhu2018evflownet}, which set time intervals to $dt=1$ and $dt=4$ images. Generally, $\textit{outdoor\_day\_2}$ was used for training, while $\textit{outdoor\_day\_1}$ and three indoor sequences formed the test set. DSEC-Flow covers a broader range of driving scenarios, including challenging conditions such as nighttime, sunrise, sunset, and tunnels. The dataset provides an official training set and a public online benchmark. The test set of DSEC-Flow also contains various challenging scenes, making it a widely used benchmark for evaluating event-based algorithms. In line with previous studies, we submitted our results to its public benchmark for evaluation.

\vspace{0.5em}
\textbf{Metrics.}
The primary accuracy metric for both DSEC-Flow and MVSEC is End-Point-Error (EPE). DSEC-Flow additionally reports Angular Error (AE) and the percentage of EPE exceeding $N$ pixels (NPE). For MVSEC, we followed prior work to compute outliers, defined as predictions with an EPE greater than 3 pixels or 5\% of the ground truth.

\vspace{0.5em}
\revise{\textbf{Train Details.}}
The proposed method was implemented using PyTorch. For event voxelization, following previous methods~\cite{gehrig2021eraft,liu2023tma}, events between two frames were divided into 15 bins. The number of targets $N$ was set to 5, resulting in $\Delta t =0.2(T_{k+1}-T_{k})$. The voxel normalization parameter $\epsilon$ in ICE was set to 0.1. The radius for lookup was set to 4. The decay factor $\gamma$ was set to 0.85. During training, the batch size was set to 6 and the learning rate was 0.0002. The number of iterations was fixed at 6 for both training and testing. The network was trained on DSEC for 200k steps and MVSEC for 100k steps.

\subsection{Results of the DSEC-Flow Dataset}

\begin{figure*}[t]
  \centering
  \includegraphics[width=0.89\linewidth]{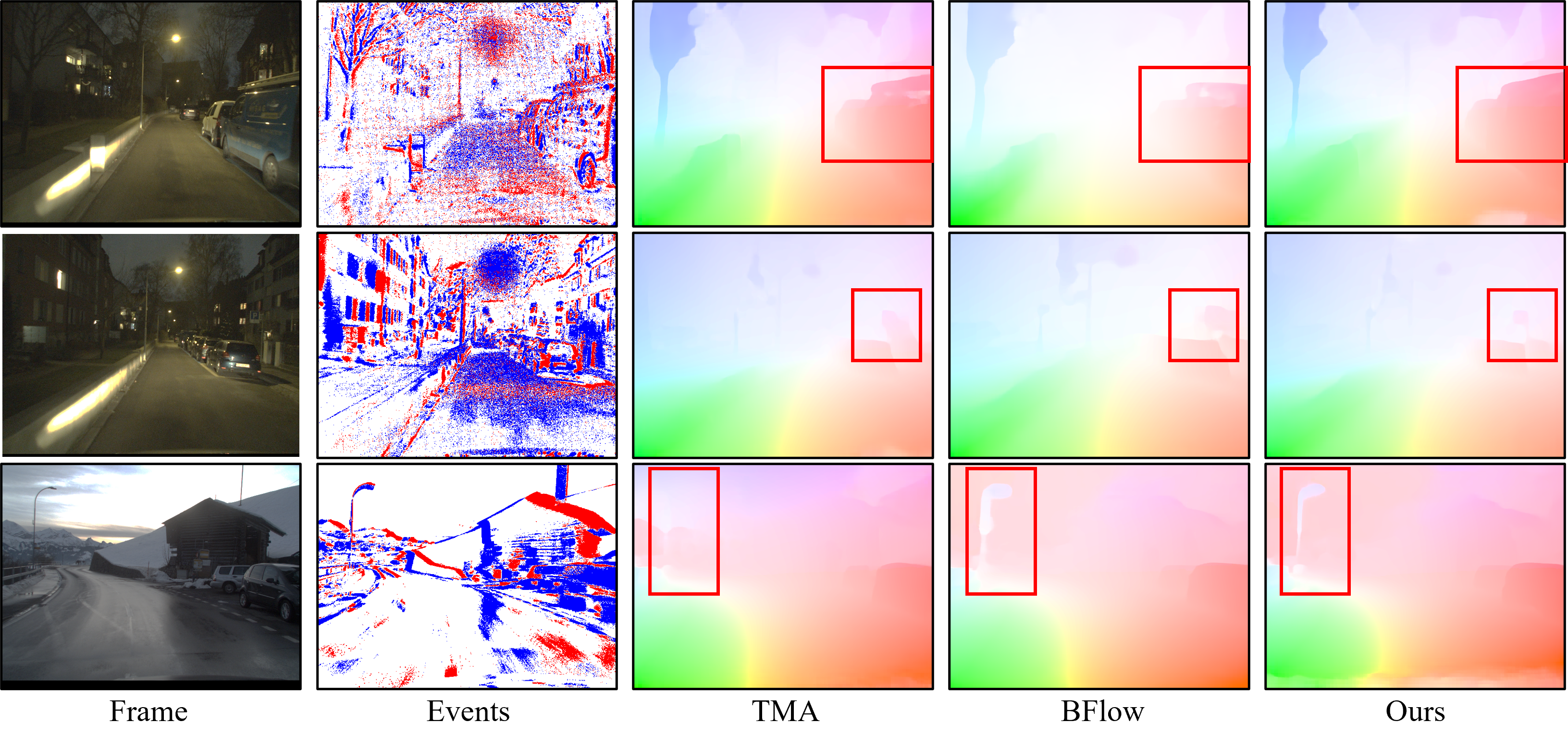}
  \caption{\textbf{Qualitative results on DSEC-Flow.} The night and daytime sequences are from \textit{zurich\_city\_12\_a} and \textit{interlaken\_01\_a}, respectively. Our method outperforms both the event-only TMA~\cite{liu2023tma} and the best fusion-based method BFlow~\cite{gehrig2024bflow}.}
  \label{fig:mainresult}
\end{figure*}

\vspace{0.5em}
\textbf{Accuracy.} 
Detailed per-sequence results on DSEC-Flow are shown in Table~\ref{tab:dsecall}, and additional results are provided in Table~\ref{tab:dsecsum}.
Several conclusions can be drawn from these results. 
Both cross-modal methods outperform the state-of-the-art event-based method ECDDP~\cite{yang2023ecddp}, highlighting that incorporating frame data significantly enhances prediction accuracy. 
The event-only method TMA~\cite{liu2023tma}, fusion-based method BFlow~\cite{gehrig2024bflow}, and our method are all based on temporally dense correlation volumes. 
Among these approaches, our method improves inter-modal interactions by replacing the simple concatenation mechanism in BFlow with cross-modal guided aggregation. 
Our method also shows notable improvements over BFlow in the NPE and AE metrics, maintains consistent performance across different sequences, underscoring its robustness.
Qualitative results in Figs.~\ref{fig:covercomp} and~\ref{fig:mainresult} show that the proposed cross-modal guidance strategy effectively mitigates the ambiguity of flow, leading to more stable and consistent predictions across consecutive frames. Additionally, in Fig.~\ref{tab:dsecsum}, our method outperforms both single-modal and cross-modal methods, while pretraining on synthetic data~\cite{gehrig2024bflow} further improves the performance.

\begin{table}[t]
\centering
\caption{\textbf{DSEC-Flow evaluation results}. Best results are highlighted as \colorbox{tabfirst}{\bf first}, \colorbox{tabsecond}{second}, and \colorbox{tabthird}{third}. `*' indicates pre-trained with synthetic data.}
\begin{tabular}{lcccccc}
\toprule \toprule
Method & Input & EPE $\downarrow$           & 1PE $\downarrow$           & 2PE $\downarrow$            & 3PE $\downarrow$           & AE $\downarrow$ \\ \midrule
RAFT~\cite{teed2020raft}   & I     & 0.78          & 12.40          & 4.6            & 2.61          & 2.44 \\
GMA~\cite{jiang2021gma}    & I     & 0.94          & 12.98         & 5.08           & 2.96          & 2.66 \\ \midrule
ERAFT~\cite{gehrig2021eraft}  & E     & 0.79          & 12.74         & 4.74           & 2.68          & 2.85   \\
TMA~\cite{liu2023tma}    & E     & 0.74          & 10.86         & 3.97          & 2.30          & 2.68    \\
ECDDP~\cite{yang2023ecddp}  & E     & 0.70          &\cellcolor{tabthird}8.89          &\cellcolor{tabthird}3.20          & 1.96          & 2.58      \\
IDNet~\cite{wu2024idnet}  & E     & 0.72          & 10.07         & 3.50          & 2.04          & 2.72     \\ \midrule
BFlow~\cite{gehrig2024bflow}  & E+I   & \cellcolor{tabthird}0.69          & 9.70          &3.42           &\cellcolor{tabthird}1.88          &\cellcolor{tabthird}2.42    \\
\textbf{Ours}   & E+I   & \cellcolor{tabsecond}{0.66} & \cellcolor{tabsecond}{8.58} & \cellcolor{tabsecond}{2.93} & \cellcolor{tabsecond}{1.68} & \cellcolor{tabsecond}{2.37}  \\ 
\textbf{Ours*}   & E+I   & \cellcolor{tabfirst}{0.63} & \cellcolor{tabfirst}{7.93} & \cellcolor{tabfirst}{2.61} & \cellcolor{tabfirst}{1.45} & \cellcolor{tabfirst}{2.29}  \\
\bottomrule \bottomrule
\end{tabular}
\label{tab:dsecsum}
\end{table}


\vspace{0.5em}
\noindent\textbf{Robustness Analysis.} The ICE representation integrates both events and frames, providing robust guidance across diverse scenarios. We validate the robustness of our approach by evaluating it on the DSEC-Flow test set, which contains challenging scenes. The detailed evaluation results are presented in Table~\ref{tab:dsecall}. 
Additionally, Fig.~\ref{fig:mainresult} provides a qualitative comparison, showcasing the nighttime sequence \textit{zurich\_city\_12\_a} and the snowy sequence \textit{interlaken\_01\_a} as examples.


As shown in Table~\ref{tab:dsecall}, our method achieves the best accuracy across both challenging and regular scenes. 
Notably, our method surpasses the state-of-the-art event-based method ECDDP~\cite{yang2023ecddp} in nighttime scenarios, despite reduced frame quality. For instance, our method achieves a 12\% accuracy improvement over ECDDP on the \textit{zurich\_city\_12\_a} sequence. 
This performance gain likely stems from the tendency of event cameras to generate increased noise in low-light conditions, while prominent edges and illuminated areas in frames still provide valuable information. 
ICE features enhance the robustness of guiding features by effectively integrating event data. 
Another demonstration of the robustness of our method is the prediction consistency shown in Fig.~\ref{fig:covercomp}. Our guided fusion strategy ensures stable spatial features, providing a critical advantage over other cross-modal approaches.

\subsection{Results of the MVSEC Dataset}

We evaluated the generalization performance on the MVSEC dataset, as reported in Table~\ref{tab:mvsec}. In accordance with the experimental setup, all models were trained on the \textit{outdoor\_day\_2} sequence and tested on the indoor and \textit{outdoor\_day\_1} sequences. The domain gap between \textit{outdoor\_day\_1} and \textit{outdoor\_day\_2} is relatively small.

From the results, we observe that model-based and unsupervised methods generally exhibit stronger generalization. In particular, MultiCM~\cite{shiba2022multicm} achieves the best performance across all indoor sequences without any training or labeled data. This highlights the potential overfitting associated with supervised learning in domain-shift scenarios.

Interestingly, our method outperforms other supervised event-only methods on the indoor sequence but performs relatively worse on \textit{outdoor\_day\_1}. We hypothesize that this contrast reflects the regularizing effect of incorporating image-based guidance, which helps mitigate overfitting by providing complementary spatial information. These results support the effectiveness of our fusion strategy in improving generalization to unseen domains.

\begin{table*}[t]
\centering
\caption{\textbf{Evaluation results on MVSEC}. Best results are highlighted as \colorbox{tabfirst}{\bf first}, \colorbox{tabsecond}{second}, and \colorbox{tabthird}{third}. `\%Outlier' denotes the proportion of outliers to all valid pixels.}
\begin{tabular}{@{}llccccccccc@{}}
\toprule \toprule
\multirow{2}{*}{}    & \multirow{2}{*}{Method} & \multirow{2}{*}{Input} & \multicolumn{2}{c}{indoor\_flying\_1}        & \multicolumn{2}{c}{indoor\_flying\_2}        & \multicolumn{2}{c}{indoor\_flying\_3}        & \multicolumn{2}{c}{outdoor\_day\_1}       \\ \cmidrule(l){4-11} 
                     &                         &                        & EPE           & \%Outlier            & EPE           & \%Outlier            & EPE           & \%Outlier            & EPE           & \%Outlier           \\ \midrule
dt=1                 &                         &                        &               &                &               &                &               &                &               &               \\ \midrule
\multirow{3}{*}{MB}  & Nagata et al.~\cite{nagata2021optical}                & E                      & 0.62          & --             & \cellcolor{tabthird}0.93 & --             & 0.84          & --             & 0.77          & --            \\
                     & Brebion et al.~\cite{brebion2021real}                 & E                      & \cellcolor{tabsecond}0.52    & \cellcolor{tabsecond}0.10     & 0.98          & 5.50           & \cellcolor{tabsecond}0.71    & \cellcolor{tabthird}2.10  & 0.53          & 0.20          \\
                     & MultiCM(Burgers')~\cite{shiba2022multicm}                 & E                      & \cellcolor{tabfirst}0.42 & \cellcolor{tabsecond}0.10     & \cellcolor{tabfirst}0.60 & \cellcolor{tabfirst}0.59  & \cellcolor{tabfirst}0.50 & \cellcolor{tabfirst}0.28  & \cellcolor{tabthird}0.30 & \cellcolor{tabthird}0.10 \\ \midrule
\multirow{3}{*}{USL} & EV-FlowNet~\cite{zhu2019unsupervised}              & E                      & 0.58          & \cellcolor{tabfirst}0.00  & 1.02          & \cellcolor{tabthird}4.00  & 0.87          & 3.00           & 0.32          & \cellcolor{tabfirst}0.00 \\
                     & FireFlowNet~\cite{paredes2021ebir}             & E                      & 0.97          & 2.60           & 1.67          & 15.30          & 1.43          & 11.00          & 1.06          & 6.60          \\
                     & ConvGRU-EV-FlowNet~\cite{hagenaars2021nips}      & E                      & 0.60          & 0.51           & 1.17          & 8.06           & 0.93          & 5.64           & 0.47          & 0.25          \\ \midrule
\multirow{3}{*}{SSL} & Ev-FlowNet~\cite{zhu2018evflownet}              & E                      & 1.03          & 2.20           & 1.72          & 15.10          & 1.53          & 11.90          & 0.49          & 0.20          \\
                     & Spike-FlowNet~\cite{lee2020spike}           & E                      & 0.84          & --             & 1.28          & --             & 1.11          & --             & 0.49          & --            \\
                     & Ziluo et al.~\cite{ding2022spatio}                 & E                      & \cellcolor{tabthird}0.57 & \cellcolor{tabsecond}0.10     & \cellcolor{tabsecond}0.79    & \cellcolor{tabsecond}1.60     & \cellcolor{tabthird}0.72 & \cellcolor{tabsecond}1.30     & 0.42          & \cellcolor{tabfirst}0.00 \\ \midrule
\multirow{3}{*}{SL}  & ERAFT~\cite{gehrig2021eraft}                   & E                      & 1.10          & 5.72           & 1.94          & 30.79          & 1.66          & 25.20          & \cellcolor{tabfirst}0.24 & \cellcolor{tabfirst}0.00 \\
                     & TMA~\cite{liu2023tma}                     & E                      & 1.06          & 3.63           & 1.81          & 27.29          & 1.58          & 23.26          & \cellcolor{tabsecond}0.25    & \cellcolor{tabsecond}0.07    \\
                     & \textbf{Ours}                  & E+I                    & 0.89          & 2.38           & 1.55          & 18.49          & 1.19          & 13.22          & 0.32          & \cellcolor{tabfirst}0.00 \\ \midrule
dt=4                 &                         &                        &               &                &               &                &               &                &               &               \\ \midrule
MB                   & MultiCM(Burgers')~\cite{shiba2022multicm}                 & E                      & \cellcolor{tabfirst}1.69 & \cellcolor{tabfirst}12.95 & \cellcolor{tabfirst}2.49 & \cellcolor{tabsecond}26.35    & \cellcolor{tabfirst}2.06 & \cellcolor{tabfirst}19.03 & 1.25          & 9.21          \\ \midrule
\multirow{2}{*}{USL} & EV-FlowNet~\cite{zhu2019unsupervised}              & E                      & 2.18          & 24.20          & 3.85          & 46.80          & 3.18          & 47.80          & 1.30          & 9.70          \\
                     & ConvGRU-EV-FlowNet~\cite{hagenaars2021nips}      & E                      & \cellcolor{tabthird}2.16 & \cellcolor{tabthird}21.51 & 3.90          & \cellcolor{tabthird}40.72 & \cellcolor{tabthird}3.00 & \cellcolor{tabthird}29.60 & 1.69          & 12.50         \\ \midrule
\multirow{3}{*}{SSL} & Ev-FlowNet~\cite{zhu2018evflownet}              & E                      & 2.25          & 24.70          & 4.05          & 45.30          & 3.45          & 39.70          & 1.23          & 7.30          \\
                     & Spike-FlowNet~\cite{lee2020spike}           & E                      & 2.24          & --             & \cellcolor{tabthird}3.83 & --             & 3.18          & --             & 1.09          & --            \\
                     & Ziluo et al.~\cite{ding2022spatio}                   & E                      & \cellcolor{tabsecond}1.77    & \cellcolor{tabsecond}14.70    & \cellcolor{tabsecond}2.52    & \cellcolor{tabfirst}26.10 & \cellcolor{tabsecond}2.23    & \cellcolor{tabsecond}22.10    & \cellcolor{tabthird}0.99 & \cellcolor{tabthird}3.90 \\ \midrule
\multirow{3}{*}{SL}  & ERAFT~\cite{gehrig2021eraft}                   & E                      & 2.81          & 40.25          & 5.09          & 64.19          & 4.46          & 57.11          & \cellcolor{tabsecond}0.72    & \cellcolor{tabsecond}1.12    \\
                     & TMA~\cite{liu2023tma}                     & E                      & 2.43          & 29.91          & 4.32          & 52.74          & 3.60          & 42.02          & \cellcolor{tabfirst}0.70 & \cellcolor{tabfirst}1.08 \\
                     & \textbf{Ours}                  & E+I                    & 2.27          & 28.03          & 4.16          & 48.38          & 3.01          & 33.24          & 0.79          & 2.04          \\ \bottomrule \bottomrule
\end{tabular}
\label{tab:mvsec}
\end{table*}

\begin{figure}[t]
  \centering
  \includegraphics[width=0.97\linewidth]{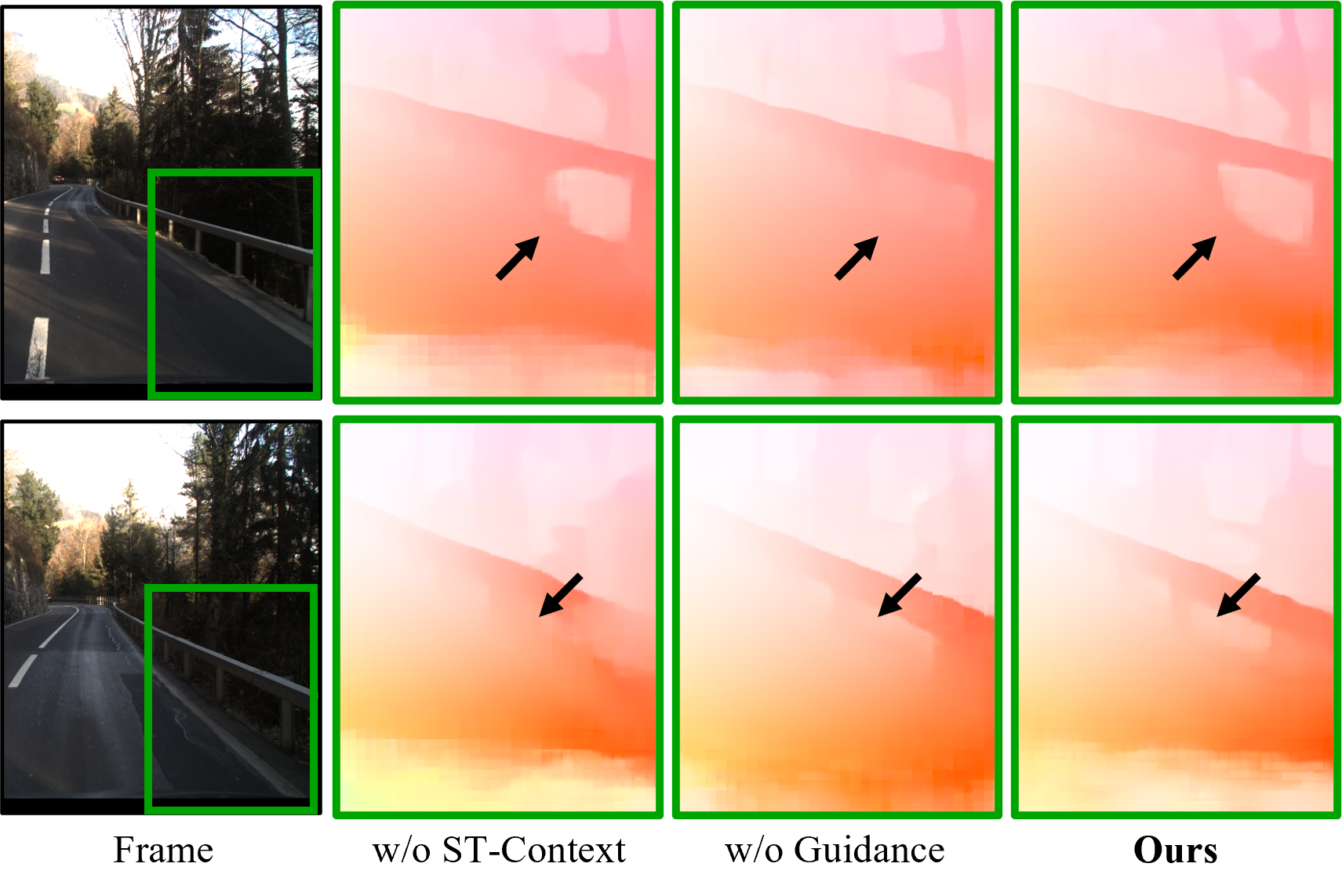}
  \caption{\textbf{Qualitative ablation studies.} Foreground–background overlap causes regional confusion. Our method mitigates these errors, highlighting the contribution of each individual component.}
  \label{fig:cmgaablation}
\end{figure}

\subsection{Ablation Study}\label{sec:exp:abla}
We conducted a series of ablation experiments to validate the proposed improvements: Enhanced Representation for Guidance, Spatiotemporal Context, and Cross-Modal Guided Aggregation. All ablation models were trained on DSEC-Flow and evaluated on the DSEC-Flow public benchmark. 

\begin{table}[t]
\centering
\caption{\textbf{Ablation study on DSEC-Flow}. Bold text highlights the contributions of this study, where `ST' refers to the spatiotemporal context feature and `GA' represents the guided aggregation strategy.}
\begin{tabular}{@{}cccc|ccc@{}}
\toprule\toprule
\multirow{2}{*}{} & \multicolumn{3}{c|}{Ablation Settings}        & \multicolumn{3}{c}{Metrics}                      \\
                  & Guidance   & Context & Fusion & EPE            & 1PE            & 3PE            \\
                \midrule
1                 & Frame        & \textbf{ST}  & \textbf{GA} & 0.67          & 8.81          & 1.70          \\
2                 & \textbf{ICE} & Frame           & \textbf{GA} & 0.69          & 9.54          & 1.79          \\
3                 & \textbf{ICE} & Events           & \textbf{GA} & 0.68          & 8.86          & 1.78          \\
4                 & \textbf{ICE} & \textbf{ST}  & \texttt{concat}       & 0.69          & 9.63          & 1.79          \\
5                 & \textbf{ICE} & \textbf{ST}  & \textbf{GA} & \textbf{0.66} & \textbf{8.58} & \textbf{1.68} \\
\bottomrule\bottomrule
\end{tabular}
\label{tab:ablation}
\end{table}

\textbf{Enhanced Representation for Guidance.} ICE was introduced to enhance the stability of frame features used for guidance. To evaluate its effectiveness, we replaced ICE with standard frames in the optimal configuration and retrained the model. As shown in Table~\ref{tab:ablation}, frame guidance leads to a decrease in accuracy compared to ICE. This conclusion aligns with our observations in Fig.~\ref{fig:ice}, indicating that ICE features exhibit more robust spatial structural information. Overall, we conclude that ICE provides more robust guidance than standard frame features.

\textbf{Spatiotemporal Context.} 
We introduce the mix-fusion context encoder to extract spatiotemporal context features, in place of single-modality features. 
To investigate the differences between different context features, we replace the spatiotemporal features with frame or event context features, respectively. 
As shown in rows 2 and 3 of Table~\ref{tab:ablation}, extracting context features solely from events or frame results in reduced accuracy, highlighting the advantage of spatiotemporal features over single-modality features. 
The qualitative comparison in Fig.~\ref{fig:cmgaablation} further validate the effectiveness.
Furthermore, comparing event context features with frame context features reveals that event features outperform frame-based features. This may be attributed to events capturing both motion cues from intermediate processes and the spatial structure of the scene.
In conclusion, the spatiotemporal context features offer more comprehensive information and compact representations.

\textbf{Cross-Modal Guided Aggregation.} 
Different with BFlow, which concatenates motion features from both modalities directly, our approach leverages the spatial stability of ICE motion features to guide the aggregation of temporally dense event motion features. 
In the ablation study, we replace guided aggregation with direct concatenation for comparison. As shown in rows 4 and 5 of Table~\ref{tab:ablation}, the guided aggregation strategy significantly improves the accuracy, confirming that enhancing inter-modal interactions improves modality fusion compared to simple concatenation.
Fig.~\ref{fig:cmgaablation} further validates this conclusion through qualitative comparisons.


\subsection{Efficiency}

\begin{figure}[t]
  \centering
  \includegraphics[width=0.97\linewidth]{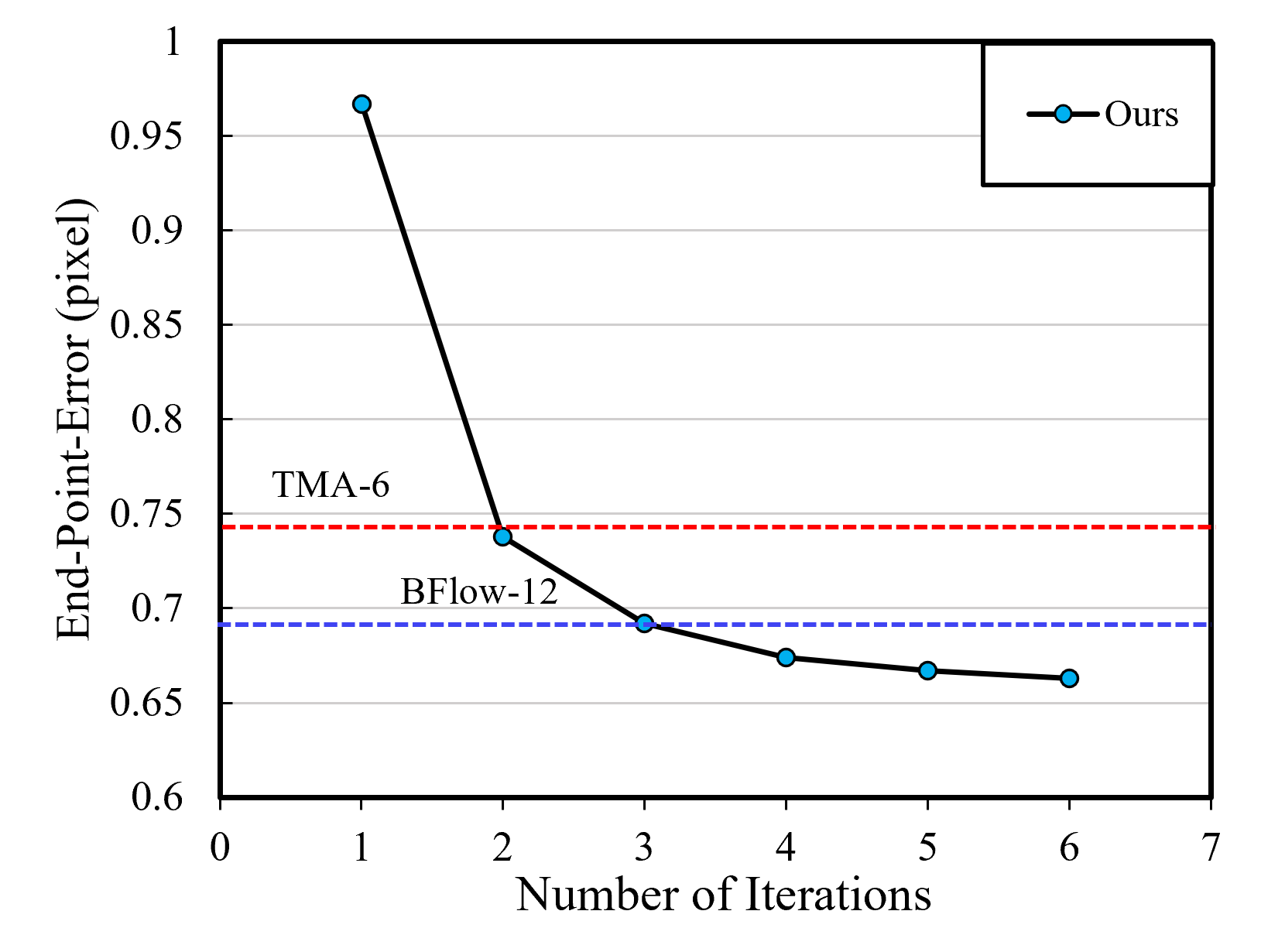}
  \caption{\textbf{EPE vs. inference iterations.} Our method achieves faster convergence, surpassing TMA-6 in 2 iterations and reaching BFlow-12 accuracy in just 3 iterations.}
  \label{fig:iteration}
\end{figure}

\begin{table}[t]
\centering
\caption{Comparison of the Computational Performance of Several Iterative Methods.}
\begin{tabular}{@{}lccccc@{}}
\toprule\toprule
\multicolumn{1}{c}{Methods} & Params & iterations&Times(ms) & Mem.[MB]&EPE  \\
\midrule
ERAFT~\cite{gehrig2021eraft}            & 5.3M  & 12  &74         & 975 & 0.79 \\
TMA~\cite{liu2023tma}                   & 6.9M  & 6   &43        &2185     & 0.74 \\
BFlow~\cite{gehrig2024bflow}            & 6.5M  & 12  &111      &1887     & 0.69 \\
\textbf{Ours}                                    & 9.2M  & 6   &61       &2459     & 0.66 \\
\bottomrule\bottomrule
\end{tabular}\label{tab:eff}
\end{table}


We present the iteration efficiency and computational performance of different approaches in Fig.~\ref{fig:iteration} and Table~\ref{tab:eff}.
For correlation-based methods, inference time is significantly affected by the number of iterations~\cite{deng2023findings}.
As shown in the table, inference time consistently increases with the number of iterations across different methods, revealing a clear trade-off between speed and accuracy. 
Reducing iterations accelerates inference but leads to a notable drop in accuracy. Our approach addresses this by enhancing the quality of extracted features, enabling higher accuracy with fewer iterations.
As shown in Fig.~\ref{fig:iteration}, our approach delivers superior performance in the early iterations, which surpasses the accuracy of TMA-6 at the second iteration and achieves the accuracy of BFlow-12 by the third iteration.

We further report the GPU memory consumption. As shown in Table~\ref{tab:eff}, multi-correlation volume methods (TMA, BFlow, and ours) naturally incur higher memory usage compared to single-volume baselines. Our method exhibits slightly higher memory usage than TMA, consistent with the additional frame modality, while still maintaining high inference efficiency and superior accuracy.

\subsection{Downstream Task}
\begin{figure}[t]
  \centering
  \includegraphics[width=0.99\linewidth]{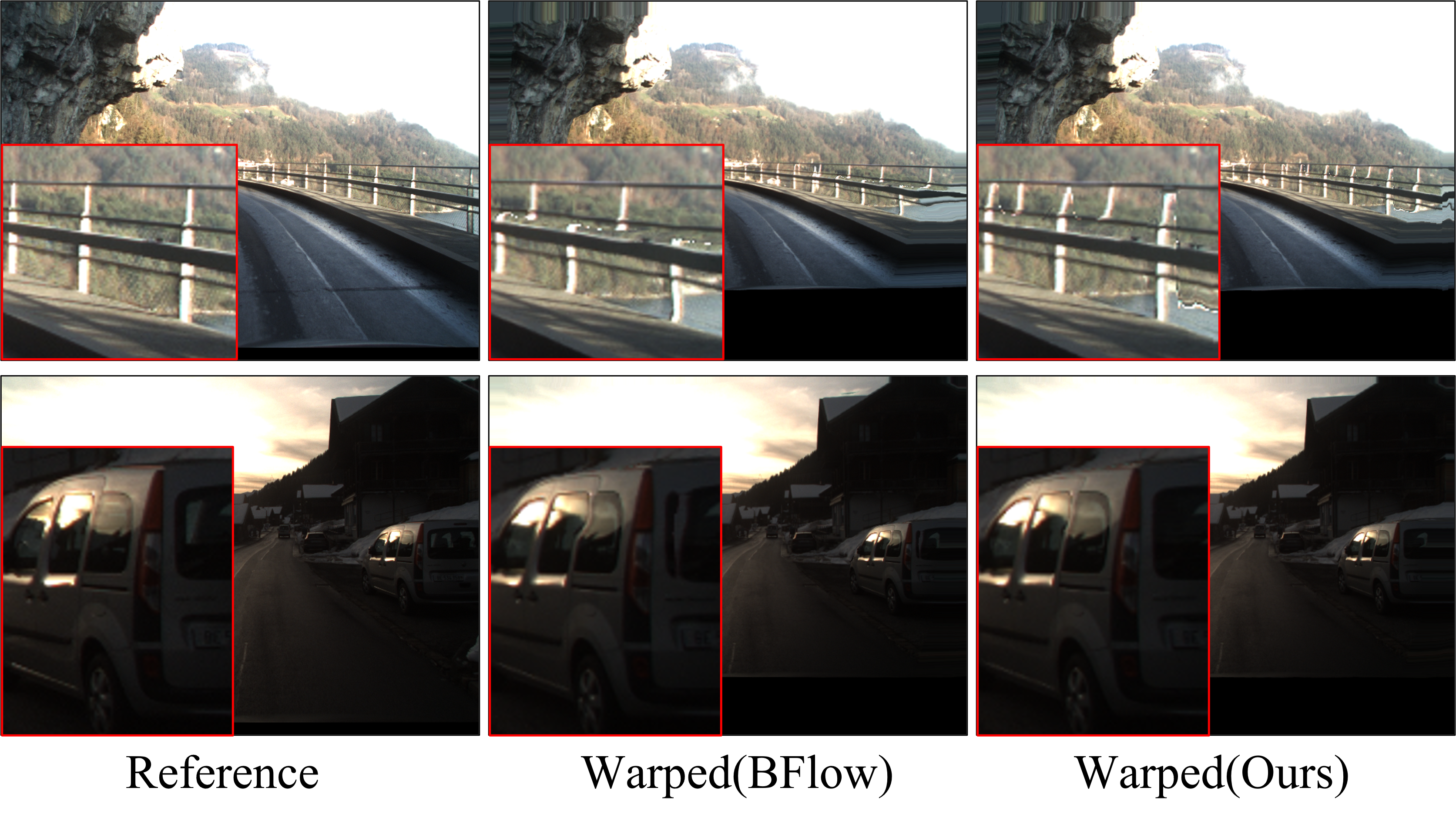}
  \caption{\revise{Qualitative comparison of warped frames obtained using different optical flow methods.}}
  \label{fig:vfi}
\end{figure}

\revise{Optical flow provides dense correspondence between frames and serves as a strong initialization for frame interpolation. 
To isolate the role of flow, we warp the end frame backward using the predicted flow and compare the result with the start frame using SSIM and PSNR. This setup allows us to indirectly assess the interpolation capability of the flow.
Quantitative results are reported in Table~\ref{tab:vfi}, and qualitative comparisons are shown in Fig.~\ref{fig:vfi}. The results demonstrate that: (i). Predicted flow enables the reconstruction of most pixels with reasonable accuracy, validating its utility in interpolation. (ii). Our method achieves higher accuracy around object boundaries and fine structures, along with improved overall similarity.}

\begin{table}[t]
\centering
\caption{\revise{Evaluation of flow accuracy by warping frame on the DSEC test. SSIM and PSNR are computed between the original and flow warped frame.}}
\begin{tabular}{@{}lcccc@{}}
\toprule
SSIM/PSNR     & inter\_01\_a   & inter\_00\_b   & thun\_01\_a & city\_12\_a \\ \midrule
BFlow         & 0.788/\textbf{25.14}         & 0.675/18.78         & 0.793/25.20 & 0.826/28.11         \\ \midrule
\textbf{Ours} & \textbf{0.792}/\textbf{25.14}         & \textbf{0.680}/\textbf{18.90}         & \textbf{0.804}/\textbf{25.53} & \textbf{0.836}/\textbf{28.52}         \\ \midrule
              & city\_15\_a & city\_14\_c & thun\_01\_b & Overall             \\ \midrule
BFlow         & 0.770/22.36         & \textbf{0.802}/23.50         & 0.782/24.10 & 0.776/24.12         \\ \midrule
\textbf{Ours} & \textbf{0.772}/\textbf{22.45}         & \textbf{0.802}/\textbf{23.52}         & \textbf{0.785}/\textbf{24.14} & \textbf{0.781}/\textbf{24.24}         \\ \bottomrule
\end{tabular}\label{tab:vfi}
\end{table}

\begin{figure}[h]
  \centering
  \includegraphics[width=0.99\linewidth]{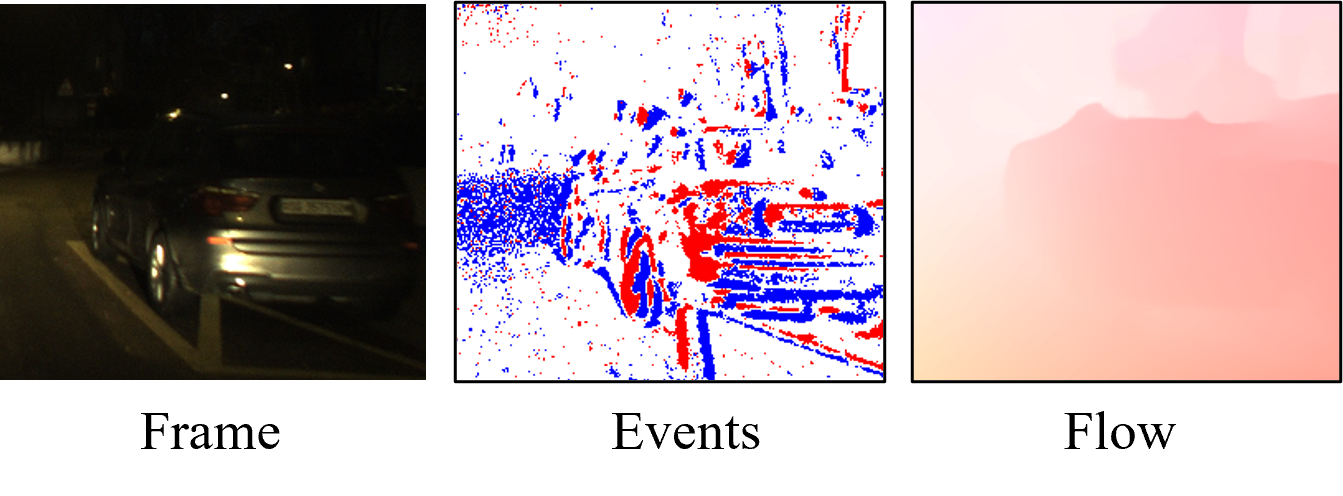}
  \caption{\revise{Failure case under extreme low-light conditions where both RGB and event inputs are severely degraded.}}
  \label{fig:fcase}
\end{figure}

\subsection{Limitations and Future Work}
\revise{While our method demonstrates strong performance across diverse scenarios, it has limitations in extreme cases where both RGB and event inputs are severely degraded. A typical failure case is observed under extreme low-light conditions, where the RGB input is nearly black and event data are sparse and noisy. As shown in Fig.~\ref{fig:fcase}, the ICE representation fails to provide spatially stable guidance, resulting in degraded flow estimation.} 
\revise{A potential improvement involves aggregating events across longer durations with motion compensation to reconstruct spatially structures, which may provide alternative guidance in the absence of reliable RGB input.}

\section{Conclusion}\label{sec:conclus}
This paper explores the fusion of event and frame data for optical flow estimation, leveraging the complementary strengths of both modalities. Our approach integrates the spatial stability of frames with the high temporal resolution of events, introducing a novel paradigm for robust optical flow estimation. Specifically, we reorganize the two modalities into a spatially robust guiding modality. The aggregation module leverages transformer architectures, where guided features are used to enrich event features and guide the temporal aggregation. The proposed strategy enhances cross-modal interactions and fully exploits the complementary strengths between events and frames. Additionally, we introduce a compact spatiotemporal context to replace single-modality contexts. 
Experimental results demonstrate that our fusion strategy achieves rapid convergence and state-of-the-art accuracy with minimal iterations. Comprehensive ablation studies further validate the robustness and effectiveness of our method.

\bibliographystyle{unsrt}
\bibliography{ref}

\end{document}